\documentclass[10pt,twocolumn,letterpaper]{article}

\usepackage[pagenumbers]{cvpr}   

\usepackage[accsupp]{axessibility} 
\usepackage{indentfirst}
\usepackage{capt-of}
\usepackage[dvipsnames]{xcolor}
\usepackage{xspace}
\usepackage{pifont}
\usepackage{enumitem}
\usepackage{makecell}
\usepackage{amsmath,amssymb,amsbsy,amsfonts,dsfont,pifont,bm,bbm,mathrsfs,mathtools,nicefrac}
\usepackage{algorithm,algpseudocode,listings}
\usepackage{booktabs,multirow,adjustbox,diagbox,threeparttable,tabularray}
\usepackage{colortbl}
\usepackage{amsmath}

\newcommand{\todo}[1]{{13689}}
\newcommand{\tocite}[1]{{\color{red} [TO CITE]}}

\definecolor{cvprblue}{rgb}{0.21,0.49,0.74}
\usepackage[pagebackref,breaklinks,colorlinks,citecolor=cvprblue]{hyperref}

\usepackage{wrapfig}
\usepackage[capitalize]{cleveref}  
\crefname{section}{Sec.}{Secs.}
\Crefname{section}{Section}{Sections}
\crefname{table}{Tab.}{Tabs.}
\Crefname{table}{Table}{Tables}
\crefname{figure}{Fig.}{Figs.}
\Crefname{figure}{Figure}{Figures}
\crefname{equation}{Eq.}{Eqs.}
\Crefname{equation}{Equation}{Equations}

\def\paperID{\todo{}}
\def\confName{CVPR}
\def\confYear{2024}

\newcommand\blfootnote[1]{%
  \begingroup
  \renewcommand\thefootnote{}\footnote{#1}%
  \addtocounter{footnote}{-1}%
  \endgroup
}

\title{FreeCustom: Tuning-Free Customized Image Generation \\ for Multi-Concept Composition}

\author{
Ganggui Ding, ~~Canyu Zhao, ~~Wen Wang, ~~Zhen Yang, ~~Zide Liu, ~~Hao Chen, ~~Chunhua Shen\\[0.2cm]
Zhejiang University, China
}

\begin{document}

\twocolumn[{
\renewcommand\twocolumn[1][]{#1}
\maketitle
\begin{center}
    \centering
    \vspace*{-.8cm}
    \includegraphics[width=\textwidth]{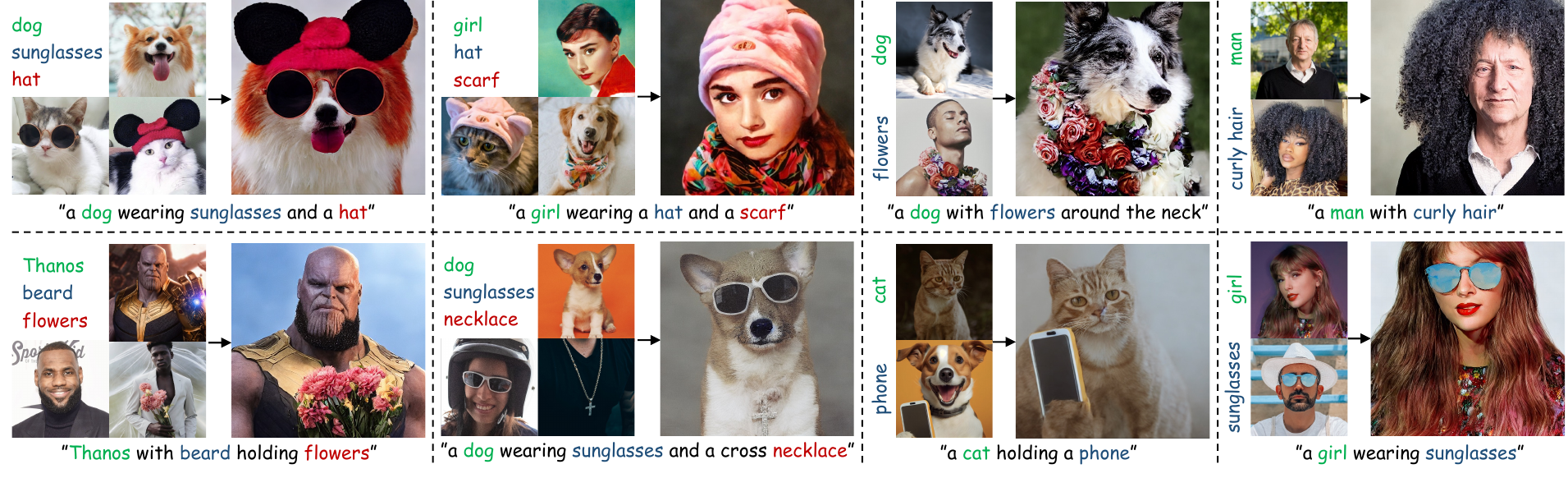}
    \vspace*{-.6cm}
    \captionof{figure}{
    \textbf{Results of customized multi-concept composition.} Our method excels at \textit{rapidly} generating high-quality images with multiple concept combinations, without any model parameter tuning. The identity of each concept is remarkably preserved. Furthermore, our method exhibits great versatility and robustness when dealing with different categories of concepts. This versatility allows users to generate customized images that involve diverse combinations of concepts, catering to their specific needs and preferences.
    Best viewed on screen.
    }
\label{fig:results_of_multi_concept}
\end{center}
}]

\maketitle
\blfootnote{$^*$GD, CZ, and WW contributed equally. Correspondence should be addressed to HC and CS.}

\begin{abstract}

Benefiting from large-scale pre-trained text-to-image (T2I) generative models, impressive progress has been achieved in customized image generation, which aims to generate user-specified concepts.
Existing approaches have extensively focused on single-concept customization and still encounter challenges when it comes to complex scenarios that involve combining multiple concepts. These approaches often require retraining/fine-tuning using a few images, leading to time-consuming training processes and impeding their swift implementation. 
Furthermore, the reliance on multiple images to represent a singular concept increases the difficulty of customization.

To this end, we propose \textbf{FreeCustom}, a novel tuning-free method to generate customized images of multi-concept composition based on reference concepts, using only one image per concept as input. Specifically, we introduce a new multi-reference self-attention (MRSA) mechanism and a weighted mask strategy that enables the generated image to access and focus more on the reference concepts. In addition, MRSA leverages our key finding that input concepts are better preserved when providing images with context interactions.
Experiments show that our method's produced images are consistent with the given concepts and better aligned with the input text. 
Our method outperforms or performs on par with other training-based methods in terms of multi-concept composition and single-concept customization, but is simpler. Codes can be found \href{https://github.com/aim-uofa/FreeCustom}{here}.

\end{abstract}
\section{Introduction}\label{sec:intro}

With the joint scaling of data, computational resources, and model size, large-scale pre-trained diffusion models~\cite{dhariwal2021diffusion, ho2020denoising, nichol2021improved, song2020denoising, rombach2022high} have made unprecedented progress in text-to-image generation. Benefiting from the capacity of pre-trained models, customized generation, \textit{i.e.}, generating user-specified objects, becomes possible and has achieved increasing attention due to its wide applications, such as advertisement production, virtual try-on, and art creation. 

Existing methods for customization, such as DreamBooth~\cite{ruiz2023dreambooth}, Textual Inversion~\cite{gal2022image}, and BLIP Diffusion~\cite{li2024blip} have shown significant progress in single-concept customization~\cite{10.1145/3618322}, however, they encounter difficulties when dealing with more complex scenarios involving multiple concepts~\cite{kumari2023multi, tewel2023keylocked}. In such cases, these methods are prone to overfitting and demonstrate poor performance in maintaining image naturalness and preserving subject identities when combining multiple concepts. Additionally, as shown in Fig.~\ref{fig:paradigm_comparison}, both approaches necessitate time-consuming training processes, either requiring fine-tuning using 3-5 images or retraining on large-scale datasets, further complicating their practical applicability.

To address these limitations, we propose FreeCustom, a tuning-free method for customized image generation that allows multi-concept composition using only one image per concept as input. Specifically, we employ a dual-path architecture to extract and combine features of input multiple concepts. Then, we introduce a novel multi-reference self-attention (MRSA) mechanism, which extends the original self-attention to access and query features of reference concepts. To highlight the input concepts and eliminate irrelevant information, we implement a weighted mask strategy that directs MRSA to focus more on the given concepts. In addition, we find that the context interaction of the input concept is of great importance for multi-concept composition, and the MRSA mechanism effectively captures the global context of the input images. Consequently, our proposed method can rapidly generate high-fidelity images that align precisely with the text and maintain consistency with reference concepts, without the need for training.

\begin{figure}[t]
  \centering
  \includegraphics[width=1\linewidth]{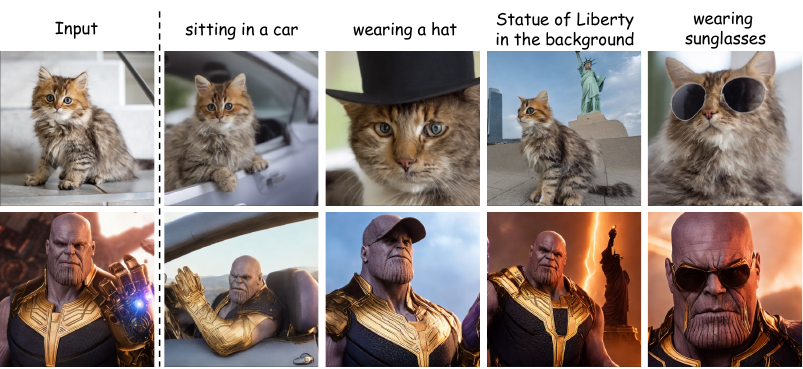}
  \caption{
  Results of single-concept customization.
  }
  \label{fig:results_of_single_concept}
\end{figure}

Extensive experiments have shown that our method achieves comparable results to other methods in terms of single-concept customization and exhibits noteworthy advantages when combining multiple concepts. It robustly and effectively generates high-quality images across diverse concepts, as shown in Figs.~\ref{fig:results_of_multi_concept} and ~\ref{fig:results_of_single_concept}. Furthermore, it can be easily applied to other diffusion-based models.

To summarize, our contributions are listed as follows:
\begin{itemize}
    \item We present FreeCustom, a novel tuning-free method that consistently delivers high-quality results for single-concept customization and multi-concept composition. 
    \item We propose the MRSA mechanism and a weighted mask strategy, allowing the generated image to interact with and focus more on the input concepts.
    \item We pay attention to the significance of the context interaction and leverage it to generate high-fidelity customized images.
\end{itemize}
\section{Related Work}\label{sec:related}

\begin{figure}[t]
  \centering
  \includegraphics[width=1\linewidth]{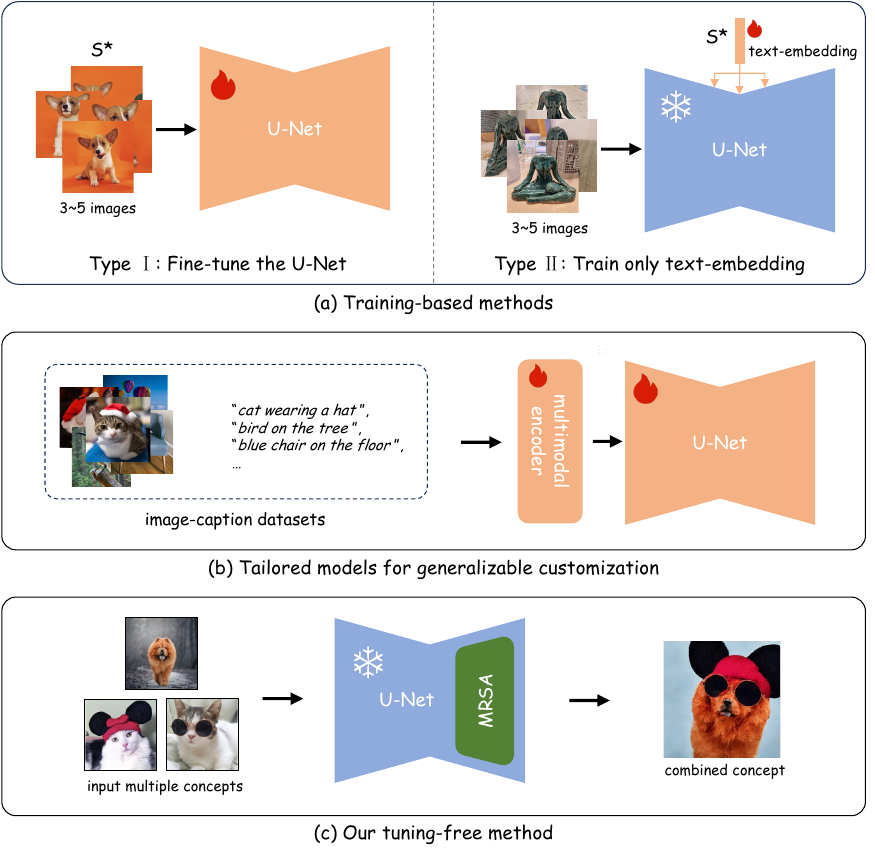}
  \caption{
  \textbf{Paradigm comparison.} Previous methods for customization can be categorized into two main categories: (a) training-based methods and (b) tailored models for generalizable customization. Training-based methods often involve fine-tuning an entire model (Type \uppercase\expandafter{\romannumeral1}) or learning a text embedding to represent a specific subject (Type \uppercase\expandafter{\romannumeral2}). Tailored models typically require re-training on large-scale image datasets to establish a versatile foundation. Unlike these two types of methods, our approach can directly generate customized images of multi-concept combinations without any additional training.
  }
  \label{fig:paradigm_comparison}
\end{figure}

The remarkable success of large-scale pre-trained text-to-image (T2I) diffusion models~\cite{balaji2022ediff, rombach2022high, ramesh2022hierarchical, DBLP:conf/icml/NicholDRSMMSC22, nichol2021improved, chen2022distribution, dhariwal2021diffusion} inspires the domain of customized image generation~\cite{tewel2023keylocked, 10.1145/3618322, kawar2023imagic, Cao_2023_ICCV, gal2023encoder, chen2023anydoor}.
The goal is to generate new images of a subject of interest, specified by one or a few user-provided images, with varying poses and locations. These methods can be divided into two categories: training-based customization and tailored models for generalizable customization.

\noindent\textbf{Training-based customization.} In the training-based personalization method, a pioneer work DreamBooth~\cite{ruiz2023dreambooth} fine-tunes the T2I diffusion model to bind a unique identifier with the subject of interest. It proposes a prior preservation loss to alleviate overfitting during few-shot tuning. A current work Textual Inversion~\cite{gal2022image} finds that a subject can be represented by a simple learnable text embedding. However, the capacity of a single learnable text embedding is limited, thus subsequent XTI~\cite{voynov2023p+} and NeTI~\cite{10.1145/3618322} introduce layer-wise learnable embedding or implicit time-aware representation, to achieve better performance. 

\noindent\textbf{Tailored models for generalizable customization.}
Another line of works~\cite{li2024blip, ruiz2023hyperdreambooth, gal2023encoder, DBLP:conf/iccv/WeiZJB0Z23, Liu2023ConesCN, chen2023improving} attempt to realize faster customization, they train a multimodal encoder and a text-to-image model on dataset-scale images, leading to tailored models for customized generation. In this way, they greatly reduce the number of fine-tuning steps required for customization. It only takes dozens to a hundred steps to achieve comparable outcomes as other methods, which require thousands of fine-tuning steps. For example, BLIP-Diffusion~\cite{li2024blip} is trained on a subset of OpenImage V6~\cite{long2022towards, long2023icdar}, and generates customized images from BLIP-2~\cite{li2023blip} encoded subject representation. BLIP-Diffusion even supports customized images generation in a zero-shot manner, but the effect is diminished. Similarly, HyperDreamBooth~\cite{ruiz2023hyperdreambooth} is trained on a massive dataset of the low-rank (LoRA)~\cite{hu2022lora} weights for the customized subject and predicts the LoRA weights for the subject of interest during inference.
These methods still cannot eliminate the need for fine-tuning, and the requirement for large-scale training further restricts their widespread application. For instance, our method supports direct application to various T2I basic models, However, BLIP-Diffusion requires large-scale re-training for each basic model.

\noindent\textbf{Multi-concept composition.}
Different from these works that perform single object customization, some works~\cite{avrahami2023bas, kumari2023multi, tewel2023keylocked} focus on multi-subject customization. Custom Diffusion~\cite{kumari2023multi} realizes it with a closed-form constrained optimization.
Perfusion~\cite{tewel2023keylocked} learns text tokens and modifies the cross-attention of the T2I model, which supports multi-concept composition.
Cones series~\cite{Liu2023ConesCN, liu2023cones2} find that a small cluster of neurons in the pre-trained diffusion model is connected to a subject, and further introduce layout guidance to achieve multi-subject customization.
Similarly, Mix-of-show~\cite{gu2024mix} introduces gradient fusion to merge several single concepts and alleviate concept conflicts among objects with regional sampling.

While progress has been made, these methods rely on a significant amount of time for few-shot or even dataset-scale training, before sampling for customized generation. What's more, they struggle with identity preservation in the one-shot customization scenario. By contrast, we propose a completely tuning-free method for customized generation, while achieving superior performance with fast sample speed. The difference between these methods and our proposed method is shown in Fig.~\ref{fig:paradigm_comparison}, more details can be seen in the Appendix.

\begin{figure*}[h]
  \centering
  \includegraphics[width=1\linewidth]{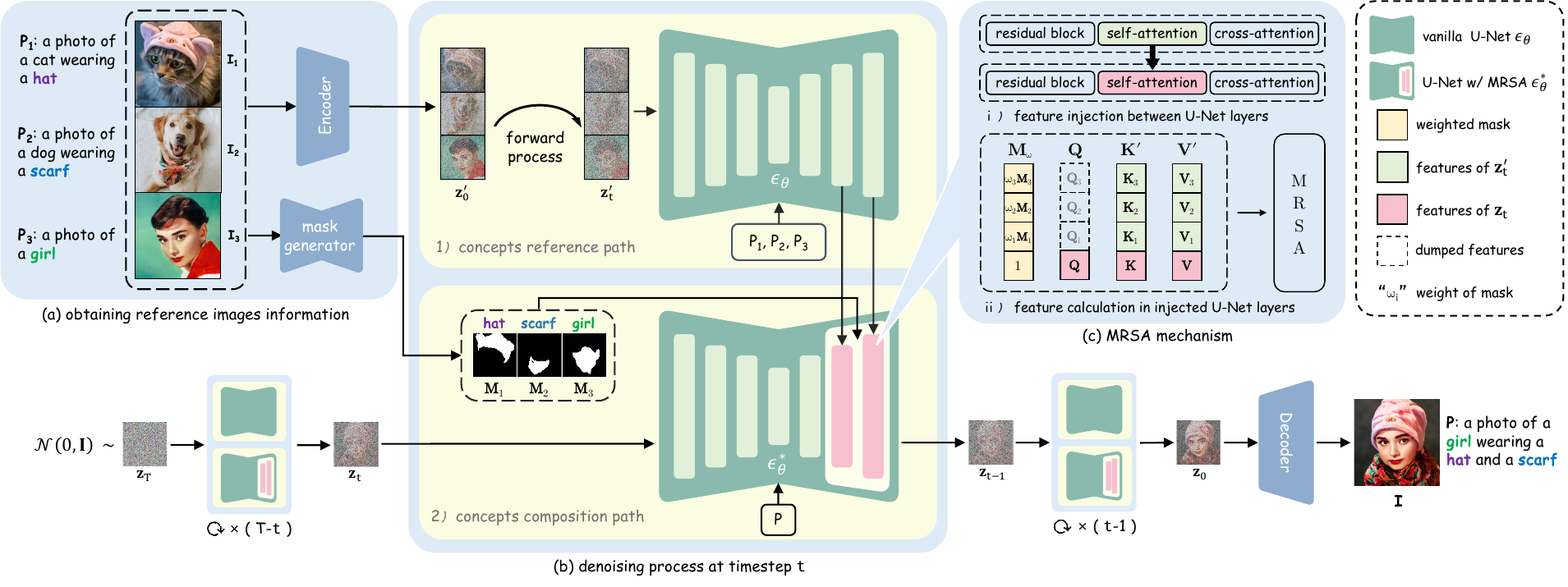}
  \caption{
  \textbf{Overview of the pipeline.} Given a set of reference images $\mathcal{I} = \{I_1, I_2, I_3\}$ and their corresponding prompts $\mathcal{P} = \{P_1, P_2, P_3\}$, we generate a multi-concept customized composition image $I$ aligned to the target prompt $P$. 
  (a) We use a VAE encoder to convert reference images into the latent representation $\mathbf z_0'$ and a segmentation network to extract masks of the concepts. (b) The denoising process involves two paths: \emph{1)} the concepts reference path and \emph{2)} the concepts composition path. In \emph{1)}, we employ a diffusion forward process to transform $\mathbf z_0'$ into $\mathbf z_t'$, subsequently passing $\mathbf z_t'$ to the U-Net $\epsilon_\theta$. Notably, the output of $\epsilon_\theta$ isn't used. In \emph{2)}, we initially sample $\mathbf z_T \sim \mathcal{N} (0,\textbf{I})$ and iteratively denoise the latent until we obtain $\mathbf z_0$. At each time step \emph{t}, we directly transmit the current latent $\mathbf z_t$ to the modified U-Net $\epsilon_\theta^*$ and employ the MRSA to integrate the features from the last two blocks of both the U-Net $\epsilon_\theta$ and the U-Net $\epsilon_\theta^*$. Finally, we utilize a VAE decoder to convert $\mathbf{z_0}$ into the final image $I$. (c) The MRSA mechanism. \emph{\lowercase\expandafter{\romannumeral1})} Feature injection happens in the self-attention module between U-Net layers, \emph{\lowercase\expandafter{\romannumeral2})} we apply MRSA using Eq.~\eqref{eq: MRSA with weight mask}.
  }
  \label{fig:method_overview}
\end{figure*}

\section{Preliminaries}\label{sec:preliminaries}

\noindent
{\bf Latent Diffusion Models.} Diffusion models~\cite{dhariwal2021diffusion, ho2020denoising, nichol2021improved, song2020denoising} employ the diffusion process within the image space, while Latent Diffusion Models (LDMs)~\cite{rombach2022high} excel by performing it within the latent space of a pre-trained Variational Autoencoder (VAE).

LDMs include a forward and a reverse process. The forward process adds a degree of Gaussian noise to the sample $\mathbf x_0$ to directly obtain $\mathbf x_t$:
$q (\mathbf x_t | \mathbf x_0) = \mathcal N (\mathbf x_t; \sqrt{\bar\alpha}_t\mathbf x_0, (1-\bar\alpha_t) \mathbf I)$,
where $\mathbf x_t$ is the noised sample at time step $t$, $\bar{\alpha}_t = \Pi_{i=1}^t \alpha_i$, $\alpha_i$ represents the predefined noise schedule. While the reverse process is a denoising process, which iteratively removes the added noise to the $\mathbf x_t$ until we obtain $\mathbf x_0$:
$p_\theta (\mathbf x_{t-1}|\mathbf x_t) = \mathcal N (\mathbf x_{t-1}; \mu_\theta (\mathbf x_t,t), \sigma_t)$, 
where $\mu_\theta (\mathbf x_t,t)= \frac{1}{\sqrt\alpha_t} (\mathbf x_t - \frac{1-\alpha_t}{\sqrt{1-\bar\alpha_t}}\epsilon_\theta (\mathbf x_t, t))$, $\sigma_t = \frac{1-\bar\alpha_{t-1}}{1-\bar\alpha_t} \beta_t$, $\beta_t=1-\alpha_t$, $\epsilon_\theta$ represents a neural network that is trained to predict the actual added noise $\epsilon$. The primary objective of the optimization process for training $\epsilon_\theta$ is 
$
L=E_{\mathbf x_0,\epsilon \sim N (0,\mathbf I)} \Vert \epsilon - \epsilon_\theta (\mathbf x_t, t) \Vert^2_2
$.
Moreover, it is possible to condition the text prompt $P$ on the $\epsilon_\theta$. Thus, $\epsilon_\theta$ can be formulated as $\epsilon_\theta (\mathbf x_t, t, P)$. By doing so, we can establish a text-to-image (T2I) diffusion model capable of generating images based on the provided prompt.

In this study, we adopt the Stable Diffusion (SD) model as the base model. The SD model belongs to LDMs and is a state-of-the-art (SOTA) T2I generation model based on U-Net architecture, which can generate high-quality images and show significant consistency with the given text prompt.

\noindent
{\bf Vanilla self-attention in Stable Diffusion.}
The original SD U-Net $\epsilon_{\theta}$ includes an encoder and a decoder, comprising a total of 7 basic blocks and 16 layers. Each layer contains a residual block, a self-attention module, and a cross-attention module. The self-attention (SA) in each layer is responsible for generating the layout and specific details of the content. At the same time, cross-attention utilizes text embedding to guide the model's generation process. 
The output of self-attention is formulated as
\begin{equation}
\begin{aligned}
{\rm SA} (\mathbf{Q},\mathbf{K},\mathbf{V}) = \operatorname{Softmax} (\frac{\mathbf{Q}\mathbf{K}^T}{\sqrt{d}}) \mathbf{V},
\end{aligned}
\end{equation}
where $\mathbf{Q}$, $\mathbf{K}$, and $\mathbf{V}$ are the query, key, and value features projected from self-attention layers with different projection matrices.

Previous work such as P2P~\cite{hertz2022prompt}, PnP~\cite{tumanyan2023plug}, MasaCtrl~\cite{Cao_2023_ICCV}, and Tune-A-Video~\cite{wu2023tune} find the overall layout of the generated image is dominated by the query features, while the key and value features control the semantic contents, and the self-attention module in the U-Net architecture inherently supports a plug-and-play feature injection approach. Therefore, we employ a dual-path paradigm and modify the original self-attention into a meticulously crafted MRSA mechanism to obtain the input concepts' features.

\section{Method}\label{sec:method}

We aim to generate customized images with multi-concept composition in a tuning-free manner. We employ the MRSA mechanism to integrate features of each reference concepts (Sec.~\ref{method:MRSA}), then implement a weighted mask strategy to emphasize the reference concept (Sec.~\ref{method:weighted_mask}), and selectively replace the original self-attention module with MRSA (Sec.~\ref{method:seletive_MRSA_replacement}), and finally find the importance of providing images incorporate context interactions. (Sec.~\ref{method:context}).

\subsection{Multi-Reference Self-Attention}\label{method:MRSA}

The overall pipeline is a dual-path paradigm as illustrated in Fig.~\ref{fig:method_overview}. Given a set of $N$ images containing reference concepts $\mathcal{I}=\{I_1, I_2, \dots, I_N\}$ and their corresponding prompts $\mathcal{P} = \{P_1, P_2, \ldots, P_N\}$, our goal is to generate a customized image $I$ that combines multiple concepts and aligns with the target prompt $P$.

\paragraph{Obtaining reference images information.}
As depicted in Fig.~\ref{fig:method_overview} (a), we initially obtain information of  $\mathcal I$ for further processing. To begin, we employ the VAE encoder, denoted as $\operatorname{Enc}$, to transform $\mathcal{I}$ from the image space into the latent space, resulting in $\mathbf z_0' = \operatorname{Enc} (\mathcal{I})$. Subsequently, we utilize an existing segmentation model $\Phi$ to extract masks $\mathcal{M} = \{\mathbf M_1,\mathbf M_2, \dots, \mathbf M_N\}$ corresponding to the reference concepts from the input images.

\paragraph{The concepts reference path and composition path.}
Our method as a whole is a diffusion denoising process. We start by randomly sampling the latent $\mathbf z_T$ from a Gaussian distribution $\mathcal N (0,\mathbf I)$ and gradually denoise $\mathbf z_T$ until $\mathbf z_0$. The process consists of two paths: the concepts reference path and the concepts composition path, as shown in Fig.~\ref{fig:method_overview} (b).

At each time step $t$ in the denoising process of the reference path, we initially apply the conventional diffusion forward process to $\mathbf z_0'$, resulting in $\mathbf z_t$. Next, we input $\mathbf z_t'$ and $\mathcal{P}$ into the U-Net $\epsilon_\theta$ to extract the query, key, and value features denoted as $\mathcal Q_{nlt}$, $\mathcal K_{nlt}$, and $\mathcal V_{nlt}$, respectively, for the reference image $n$, attention layer $l$, and time step $t$. Note that we do not utilize the output of $\epsilon_\theta$ at each time step.

In the composition path, we modify the original U-Net $\epsilon_\theta$ by replacing the self-attention module with the MRSA module, referred to as $\epsilon_{\theta}^*$. At each time step $t$, we feed the denoised latent $\mathbf z_t$ and the target prompt $P$ into $\epsilon_{\theta}^*$ and then calculate the query, key, and value features $\mathbf Q_{lt}$, $\mathbf K_{lt}$, and $\mathbf V_{lt}$ for the generated image at attention layer $l$ and time step $t$. The final denoised $\mathbf z_0$ is transformed back to the image space, yielding the ultimate customized image $I = \operatorname{Dec} (\mathbf z_0)$, where $\operatorname{Dec}$ is the VAE decoder.

\paragraph{MSRA mechanism.}
As illustrated in Fig.~\ref{fig:method_overview} (c), features from the self-attention module in $\epsilon_{\theta}^*$ are injected into the module in $\epsilon_{\theta}$, thus, MRSA perceives not only its own features but also the features obtained from the reference images. Specifically, at each attention layer $l$ and time step $t$ of the denoising process, the $N$ injected key and value features from $\mathbf z_t'$ are denoted as $\mathcal K_{lt} = \{\mathbf K_1, \mathbf K_2, \dots, \mathbf K_N\}$ and $\mathcal V_{lt} = \{\mathbf V_1, \mathbf V_2, \dots, \mathbf V_N\}$. Meanwhile, the query, key, and value features from $\mathbf z_t$ are $\mathbf Q_{lt}$, $\mathbf K_{lt}$, and $\mathbf V_{lt}$. 
We concatenate $\mathbf K$ and $\mathbf V$ with $\mathcal K$ and $\mathcal V$ to obtain $\mathbf K' = [\mathbf K, \mathbf K_1, \mathbf K_2, \dots, \mathbf K_N]$ and $\mathbf V' = [\mathbf V, \mathbf V_1, \mathbf V_2, \dots, \mathbf V_N]$. Here, we omit the subscript $lt$ for brevity since we apply the same operation across selected layers and all time steps. Finally, the operation of MRSA is as follows:
\begin{equation}
\begin{aligned}
{\rm MRSA} (\mathbf{Q},\mathbf{K}',\mathbf{V}') &= \operatorname{Softmax} (\frac{\mathbf{Q} \mathbf{K}'^T}{\sqrt{d}}) \mathbf{V}'.
\end{aligned}
\end{equation}
This approach allows generated images to naturally incorporate features from the input images, however, we observe that it also retains unrelated features to a significant extent, resulting in concept confusion. To address this issue, we concatenate all masks $\mathbf M_i \in \mathcal M$ with an all-ones matrix $\mathbf{1}$, yielding $\mathbf M = [\mathbf 1,\mathbf M_1,\mathbf M_2,\dots,\mathbf M_N]$ and then employ $\mathbf M$ to constrain the regions attended by the MRSA, effectively masking out unrelated contents and guiding the model's focus towards the target concept.
Technically,
\begin{equation}
\begin{aligned}
\operatorname{MRSA} (\mathbf{Q},&\mathbf{K}',\mathbf{V}',\mathbf{M}) \\ &= \operatorname{Softmax} (\frac{\mathbf{M} \odot (\mathbf{Q} \mathbf{K}'^T)}{\sqrt{d}}) \mathbf{V}',
\end{aligned}
\end{equation}
$\mathbf{M}$ is downsampled here to match the resolution of the current feature, and $\odot$ represents the Hadamard product. This masked MRSA mechanism can utilize the context interaction information within the input images and filter out irrelevant features during the image generation process.

\begin{figure}[b]
  \centering
  \includegraphics[width=1\linewidth]{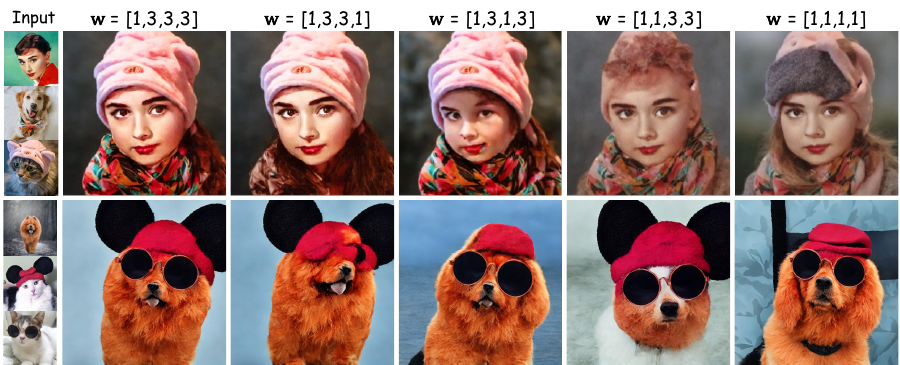}
  \caption{
  \textbf{Weighted mask strategy}. $\mathbf w$ is the weight of the mask, where the first weight corresponds to the main edited subject, and the following three weights are for the input concepts.
  }
  \label{fig:comparison_of_mask_weight_strategy}
\end{figure}

\subsection{Weighted Mask}\label{method:weighted_mask}
Although the introduction of masks helps to tackle the problem of unrelated features, the current model still struggles to accurately preserve the distinctive characteristics of the target concept in the generated results. It can only provide a rough representation of the appearance of the target concept, lacking precise details. To overcome this limitation, we introduce a scaling factor to each mask, denoted as $\mathbf w = \{1, \omega_1,\omega_2,\dots,\omega_{N}\}$, to enhance the model's focus on the target concept in conjunction with the weighted masks, {\em i.e.}, $\mathbf{M}_w=[\mathbf{1},\omega_1\mathbf{M}_1,\omega_2\mathbf{M}_2,\dots,\omega_{N}\mathbf{M}_{N}]$. Then the MRSA is formulated as
\begin{equation}
\begin{aligned}
{\rm MRSA} (\mathbf{Q},&\mathbf{K}',\mathbf{V}',\mathbf{M}_w) \\ &= \operatorname{Softmax} (\frac{\mathbf{M}_w \odot (\mathbf{Q} \mathbf{K}'^T)}{\sqrt{d}})
\mathbf{V}'.\label{eq: MRSA with weight mask}
\end{aligned}
\end{equation}
We have found that the model is not overly sensitive to the value of $\mathbf w$. Assigning values between 2 and 3 to $\mathbf w$ consistently yields excellent performance across various scenarios. By incorporating the concept-specific weighted masks within MRSA, the model is encouraged to attend more selectively to the desired features and suppress the influence of irrelevant information. This refinement leads to improved generation results that align more closely with the intended concepts as shown in Fig.~\ref{fig:comparison_of_mask_weight_strategy}.

\subsection{Selective MRSA Replacement}\label{method:seletive_MRSA_replacement}
Experimental results indicate that a straightforward substitution of the self-attention modules in all 7 basic blocks with the MRSA module results in unnatural generation, loss of conceptual coherence, and textual inconsistencies. Previous research~\cite {Cao_2023_ICCV, tumanyan2023plug} find that the query features in the deep layers of the U-Net possess the capacity for layout control and semantic information acquisition. Therefore, we adopt a similar strategy to MasaCtrl~\cite{Cao_2023_ICCV}, where the self-attention module is replaced with our MRSA module only in a selected set of blocks, particularly in the deeper blocks of the U-Net, denoted as $\Psi$. Through empirical observation, we achieve superior results by setting $\Psi = [5,6]$. These results demonstrate our ability to not only improve the natural and photorealistic appearance of the generated image but also effectively preserve the identity of the given concepts, as shown in Fig.~\ref{fig:layer_ablation}.

\begin{figure}[b]
  \centering
  \includegraphics[width=1\linewidth]{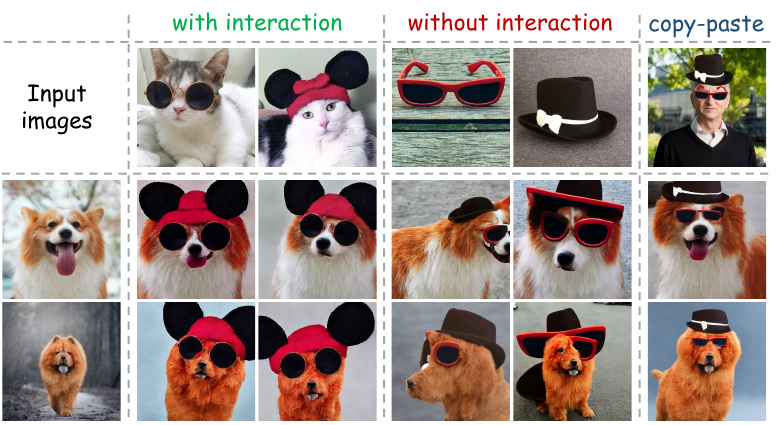}
  \caption{
  \textbf{Results of given concepts with and without context interaction.} The target prompt is ``a dog wearing a hat and sunglasses". The fidelity of the input concepts is effectively maintained when providing contextual images. On the contrary, unnatural results are generated. Manually constructing contexts by a simple copy-paste strategy also improves image quality. 
  }
  \label{fig:context_interaction_analyse}
\end{figure}

\subsection{Preparing Images with Context Interaction }\label{method:context}

We find the context interaction of each concept in an image is crucial for achieving multi-concept composition, such as providing the context of ``wearing a hat", the hat concept is well preserved. However, the generated image suffers if the input image only shows a plain hat with no indication that it is being worn. Specifically, as can be seen in Fig.~\ref{fig:context_interaction_analyse}, when only images of isolated hats and glasses without context interactions are provided, the model struggles to maintain the desired concept features successfully (columns 4 and 5). However, when images of a cat wearing sunglasses and a hat are provided, the model can generate excellent customized results (columns 2 and 3). A simple copy-paste strategy can also provide enough context information to benefit the customization (column 6). Leveraging this strategy, we can manually produce reference images with contextual interactions, helping to generate images that possess high fidelity and text alignment.

\section{Experiments}\label{sec:exp}
\paragraph{Data.}
We curate a diverse dataset comprising images sourced from the Internet and previous studies in customization~\cite{ruiz2023dreambooth, 10.1145/3618322, gal2022image} and editing~\cite{tumanyan2023plug,hertz2022prompt,vid2vid-zero,yang2023OIR}. This dataset is used to provide reference concepts for customized image generation and encompasses a wide range of categories, including faces, anime characters, animals, accessories, and wearables, with each image representing one or more distinct concepts. We employ the Grounded-Segment-Anything~\cite{kirillov2023segment, liu2023grounding} to create masks for each concept, which enables us to extract masks by simply providing the name of the desired concept, streamlining the process of obtaining the concept-specific mask. 

\paragraph{Implementation setting.}
We use Stable Diffusion V1.5\footnote{https://huggingface.co/runwayml/stable-diffusion-v1-5} as our base model to generate high-quality images of 512$\times$512 resolution using an NVIDIA 3090 GPU. The mask weight $\omega_i$ for the reference concepts is set to 3, while the mask weight for the main edited subject is set to 1 to align with the original self-attention. We assign blocks $\Psi$ as $[5,6]$ to replace the original self-attention module in the SD model with MRSA at each time step. 

\subsection{Comparison with Existing Methods}

\paragraph{Evaluation metrics.}
Similar to DreamBooth~\cite{ruiz2023dreambooth}, we access image similarity using DINOv2~\cite{oquab2023dinov2, darcet2023vitneedreg} and CLIP-I~\cite{radford2021learning}, and image-text consistency using CLIP-T~\cite{radford2021learning} and CLIP-T-L~\cite{radford2021learning}. To assess the quality of the generated images, we employ CLIP-IQA~\cite{wang2022exploring}, an evaluative criterion that considers the visual aesthetics and the abstract perception in an image. Furthermore, we evaluate the overall time required for implementing the method, including both the initial preprocessing time and the subsequent inference time. Finally, we conduct a user study for evaluation in terms of image quality, identity fidelity, and text alignment.

\paragraph{Compared methods.}
To evaluate the effectiveness of our approach in single concept customization, we compare it against several state-of-the-art methods, namely DreamBooth~\cite{ruiz2023dreambooth}, NeTI~\cite{10.1145/3618322}, and BLIP Diffusion~\cite{li2024blip}. We compare our approach against Custom Diffusion~\cite{kumari2023multi} and Perfusion~\cite{tewel2023keylocked} for evaluating multi-concept composition. 
These competitors represent a range of approaches, including tuning-based models~\cite{ruiz2023dreambooth, 10.1145/3618322, kumari2023multi, tewel2023keylocked} and models pre-trained on large-scale datasets~\cite{li2024blip}, reflecting the diversity of established customization techniques. All comparisons are conducted using Stable Diffusion V1.5 as the common foundation, ensuring a consistent inference setup for fair evaluations among the different methods.

\begin{figure}[h]
  \centering
  \includegraphics[width=1\linewidth]{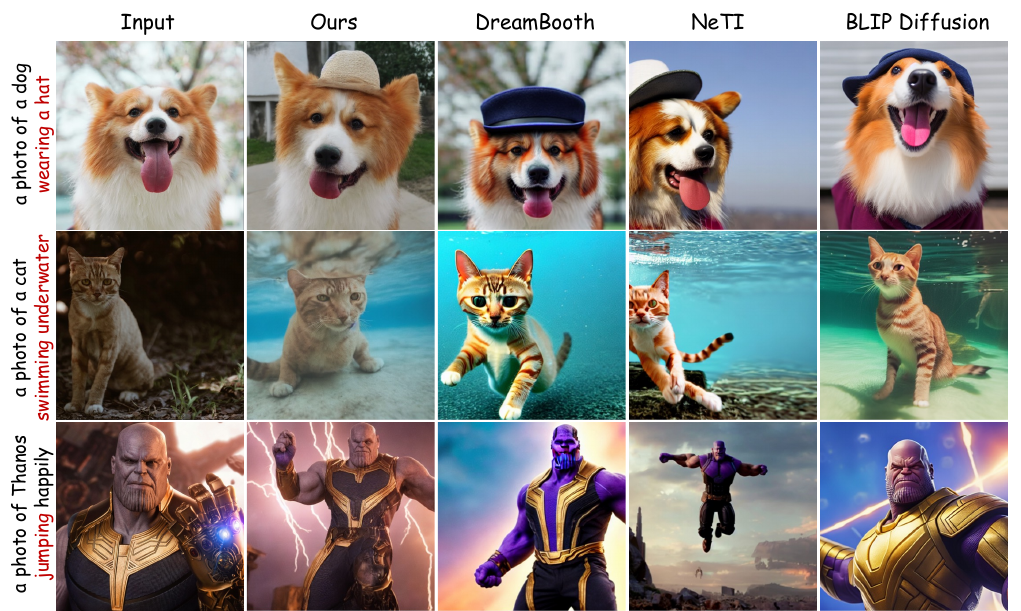}
  \caption{
  Comparisons of single-concept customization.
  }
  \label{fig:comparison_single}
\end{figure}

\begin{figure}[h]
  \centering
  \includegraphics[width=1\linewidth]{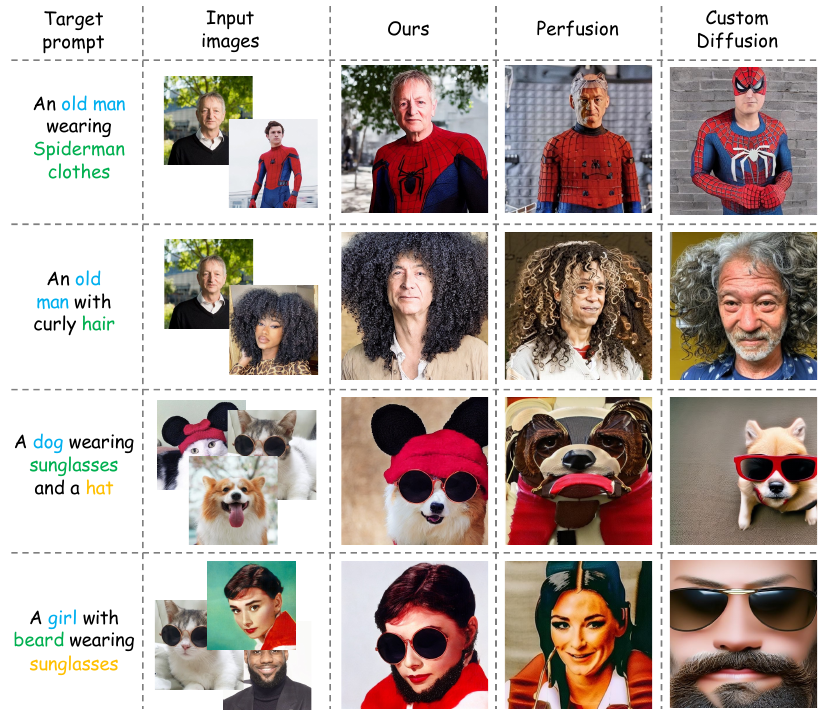}
  \caption{
  Comparisons of multi-concept composition.
  }
  \label{fig:comparison_multi}
\end{figure}

\paragraph{Qualitative comparison.}
In terms of multi-concept composition, as illustrated in Fig.~\ref{fig:comparison_multi}, our method demonstrates superior performance compared to ~\cite{kumari2023multi} and ~\cite{tewel2023keylocked}, despite being tuning-free. It efficiently preserves the identity of reference concepts in the generated customized images and performs well across a diverse range of concepts, including hairstyles, accessories, and clothing. Notably, our method exhibits significantly improved consistency between the generated images and the corresponding text prompts compared to the other methods evaluated. 
We also show the results compared with other methods in single concept customized image generation as depicted in Fig.~\ref{fig:comparison_single}. Our method achieves comparable or even more natural and realistic results, further emphasizing the high-quality image-generation capabilities of our approach.

\paragraph{Image and text fidelity.}
\begin{table}[]
\centering
\scalebox{0.7}{
        \begin{tabular}{lcccccccc}
            Methods                                         & DINOv2             & CLIP-I             & CLIP-T              & CLIP-T-L              & CLIP-IQA              \\ \toprule
            \textit{single-concept}  \\
            DreamBooth~\cite{ruiz2023dreambooth}            & \textbf{0.8948}    & \underline{0.8906} & 27.3825             & 21.9413               & 0.7194                \\
            NeTI~\cite{10.1145/3618322}                     & 0.7839             & 0.8677             & \underline{29.9023} & \underline{25.1220}   & 0.7239                \\
            BLIP Diffusion~\cite{li2024blip}                & \underline{0.8734}             & \textbf{0.8975}    & 29.1278             & 24.0543               & \textbf{0.7655}       \\
            \rowcolor{green!15}
            FreeCustom                                      & 0.8376             & 0.8755             & \textbf{32.0206}    & \textbf{27.4440}      & \underline{0.7292}    \\ \hline
            \textit{multi-concept} \\
            Custom Diffusion~\cite{kumari2023multi}         & \underline{0.6545} & \underline{0.2393} & \underline{29.0702} & \underline{23.6657}   & \underline{0.8921}   \\
            Perfusion~\cite{tewel2023keylocked}                   & 0.6399             & 0.2277             & 22.1371             & 16.1719               & 0.8624                \\
            \rowcolor{green!15}
            FreeCustom                                      & \textbf{0.7625}    & \textbf{0.2871}    & \textbf{33.7826}    & \textbf{27.8758}      & \textbf{0.9002}       \\
        \end{tabular}
}
\caption{Image similarity (DINOv2, CLIP-I), image-text alignment (CLIP-T, CLIP-T-L), and image quality (CLIP-IQA).}
\label{tab:metrics}
\end{table}
As shown in Tab.~\ref{tab:metrics}, our method ranks first in terms of 5 metrics when comparing multi-concept composition, highlighting our method's superior advantage in multi-concept results. However, when it comes to single-concept customization, our image-text alignment is strong. Nonetheless, our performance in DINOv2 and CLIP-I is mediocre due to the trade-off between image similarity and text consistency~\cite{kumari2023multi}. 

\paragraph{Time efficiency.}
\begin{table}[]
\centering
\scalebox{0.7}{
\begin{tabular}{lllll} 
{Methods}                                   & Venue        & Preprocessing & Inference & Total         \\
\toprule
\textit{single-concept}  \\
DreamBooth~\cite{ruiz2023dreambooth}        & CVPR'23      & 500s          & 3s        & 503s          \\
NeTI~\cite{10.1145/3618322}                 & SIGGRAPH'23  & 420s          & 30s       & 450s          \\ 
BLIP Diffusion~\cite{li2024blip}            & NeurIPS'23   & 6 days        & 3s        & 6 days     \\ 
\rowcolor{green!15}
FreeCustom                                  & this work    & 0             & 20s       & \textbf{20s}  \\ 
\hline
\textit{multi-concept}  \\
Custom Diffusion~\cite{kumari2023multi}     & CVPR'23      & 287s          & 13s       & 300s          \\ 
Perfusion~\cite{tewel2023keylocked}         & SIGGRAPH'23  & 821s          & 14s       & 835s          \\ 
\rowcolor{green!15}
FreeCustom (2 concepts)                     & this work    & 0             & 36s       & \textbf{36s}  \\ 
\rowcolor{green!15}
FreeCustom (3 concepts)                     & this work    & 0             & 58s       & \textbf{58s}  \\ 
\end{tabular}
}
\caption{\textbf{Comparisons of preprocessing time and inference time}. Preprocessing time includes fine-tuning or re-training time.} 
\label{tab:time_efficiency}
\end{table}
Thanks to our tuning-free paradigm, our method significantly outperforms other methods in terms of time efficiency, as reported in Tab.~\ref{tab:time_efficiency}. Fine-tuning-based models require extensive preprocessing time, whereas our method can be efficiently applied to various models without the need for additional training. On the other hand, BLIP Diffusion requires retraining when applied to different models, complicating its preprocessing.

\paragraph{User study.}
\begin{table}[]
\centering
\scalebox{0.7}{
\begin{tabular}{llccc}
{Methods}                                       & Venue         & Alignment     & Consistency & Quality\\
\toprule
\textit{single-concept} \\
DreamBooth~\cite{ruiz2023dreambooth}            & CVPR'23       & 1.77          & 1.48           & 1.53     \\
NeTI~\cite{10.1145/3618322}                     & SIGGRAPH'23   & 2.27          & 2.49           & 2.31     \\ 
BLIP Diffusion~\cite{li2024blip}                & NeurIPS'23    & 3.50          & 3.04           & 2.85     \\ 
\rowcolor{green!15}
FreeCustom                                      & this work     & \textbf{3.62} & \textbf{3.11}  & \textbf{2.89}    \\ 
\hline
\textit{multi-concept} \\
Custom Diffusion~\cite{kumari2023multi}         & CVPR'23       & 1.91          & 2.53           & 2.48    \\ 
Perfusion~\cite{tewel2023keylocked}             & SIGGRAPH'23   & 1.50          & 1.70           & 1.73    \\ 
\rowcolor{green!15}
FreeCustom                                      & this work    & \textbf{4.40} & \textbf{4.65}  & \textbf{4.17}    \\ 
\end{tabular}
}
\caption{\textbf{User study results}. The text-to-image correspondence (Alignment), identity preservation (Consistency) and image quality (Quality) are evaluated. Our approach demonstrates superior performance across all these aspects.}
\label{tab:user_study}
\end{table}
We collect a total of 23 questionnaires for single-concept and 42 questionnaires for multi-concept. Specifically, we generated results using different methods with identical prompts, base models, and image inputs. For single-concept, each method customizes 5 concepts, for a total of 20 questions. For multi-concept, each method generates 10 distinct sets of concept combinations, ranging from 2 to 4 concepts, resulting in a total of 30 questions. The results are randomly placed. Each participant is required to assess customized images based on three criteria:  alignment between image and prompt, consistency of concepts between the given image and reference images, and overall image quality. Each question is rated on a 5-point scale, with 5 representing the best. The results unequivocally demonstrate the superiority of our method in both tasks as presented in Tab.~\ref{tab:user_study}.

\begin{figure}[b]
  \centering
  \includegraphics[width=1\linewidth]{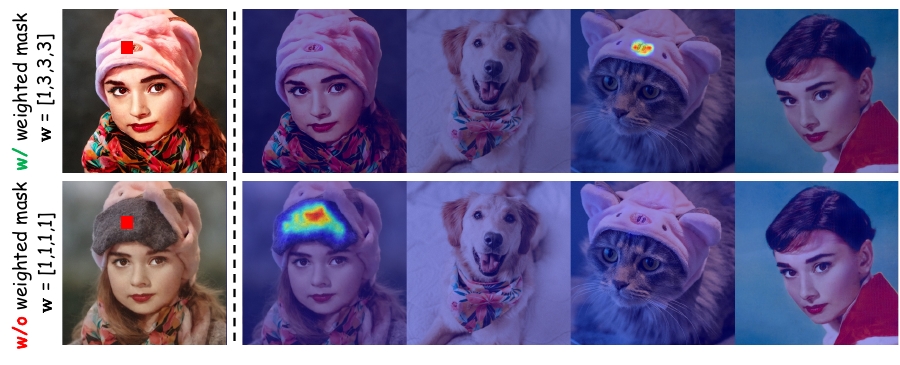}
  \caption{
  \textbf{Visualization of multi-attention map} in MRSA mechanism. We visualize the multi-attention map corresponding to the query feature indicated by the red box on the generated image. 
  }
  \label{fig:attn_map_analyse}
\end{figure}

\subsection{Ablation Studies}
\paragraph{Weighted mask strategy.}

To assess the significance of mask weights, we conduct ablation studies as depicted in Fig.~\ref{fig:comparison_of_mask_weight_strategy}. Insufficient weight assigned to a concept results in poor preservation of its identity. Additionally, we visualize the multi-attention map $A=\operatorname{Softmax} (\frac{\mathbf{M}_w \odot (\mathbf{Q} \mathbf{K}'^T)}{\sqrt{d}})$, $A\in{\mathbb{R}^{H\times ((N+1)W)}}$, $H$ and $W$ represents the resolution of features in current layer, in MRSA at the 16th layer and the 50th timestep. A weighted mask plays a critical role in enabling our model to effectively capture features from reference images, thereby facilitating accurate feature generation. In the absence of a weighted mask, attention is not adequately directed towards the input concepts. Consequently, the generated image fails to accurately capture the individual concepts portrayed in the reference image, such as the loss of the hat identity.

\begin{figure}[t]
  \centering
  \includegraphics[width=1\linewidth]{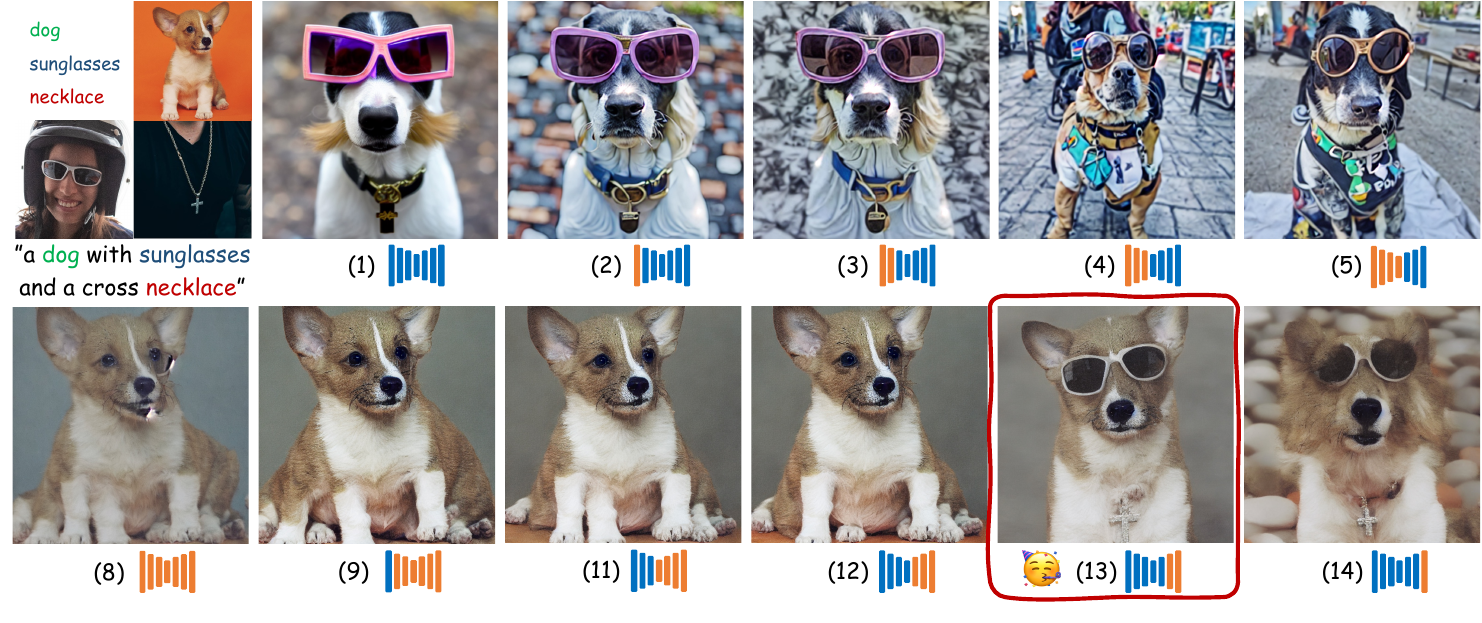}
  \caption{
  \textbf{Selective applying MRSA to basic blocks.} The blue color represents the original basic block and the yellow color indicates the basic block whose self-attention is replaced by MRSA.
  }
  \label{fig:layer_ablation}
\end{figure}

\noindent
{\bf Selective MRSA replacement.}
Instead of directly replacing the self-attention of all 7 basic blocks of U-Net with MRSA, we conduct experiments using various strategies to selectively replace the self-attention modules in the basic blocks of the U-Net with MRSA, as depicted in Fig.~\ref{fig:layer_ablation}. Results indicate that when $\Psi = [5,6]$ exhibits the highest efficacy in preserving the identity of intricate concepts, aligning with the provided prompt, and producing photo-realistic images.

\subsection{More Applications}
\begin{figure}[t]
  \centering
  \includegraphics[width=1\linewidth]{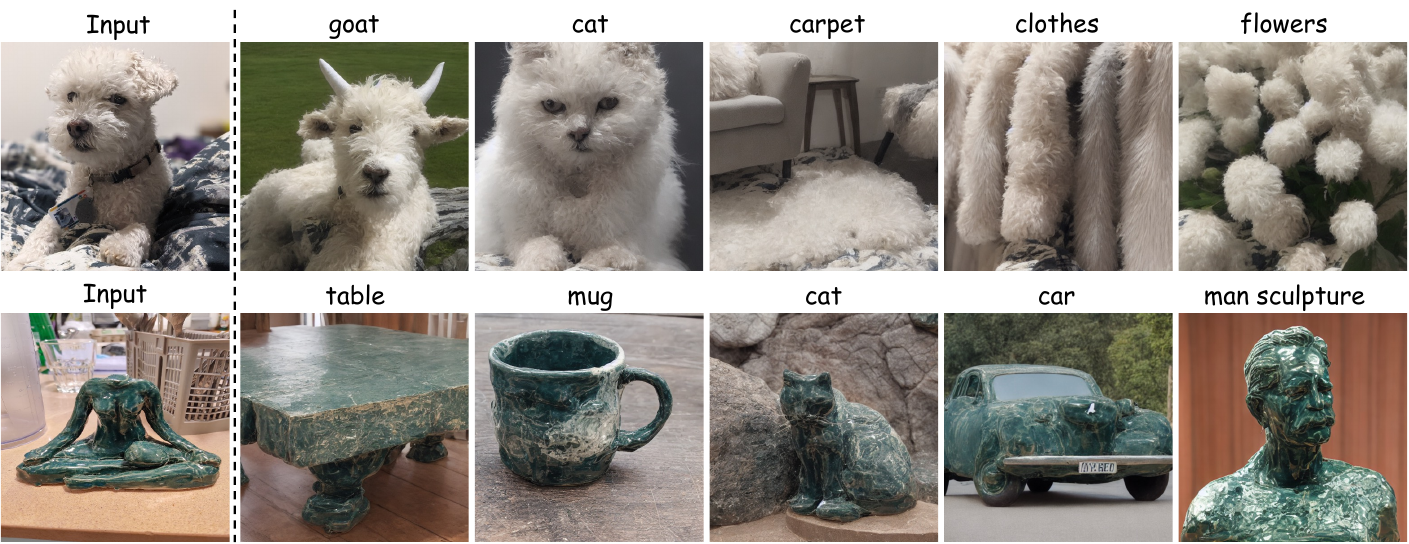}
  \caption{
  \textbf{Appearance transfer.} Our method generates objects with similar appearance and materials as the input image.
  }
  \label{fig:appearance_transfer}
\end{figure}

\noindent
{\bf Appearance transfer.}
Appearance transfer aims to convert the appearance information of the input concept to a new object. As shown in Fig.~\ref{fig:appearance_transfer}, our method benefits from the ability to query the input image and can easily implement appearance transfer, the generated image effectively retains the appearance information of the input image.

\begin{figure}[t]
  \centering
  \includegraphics[width=1\linewidth]{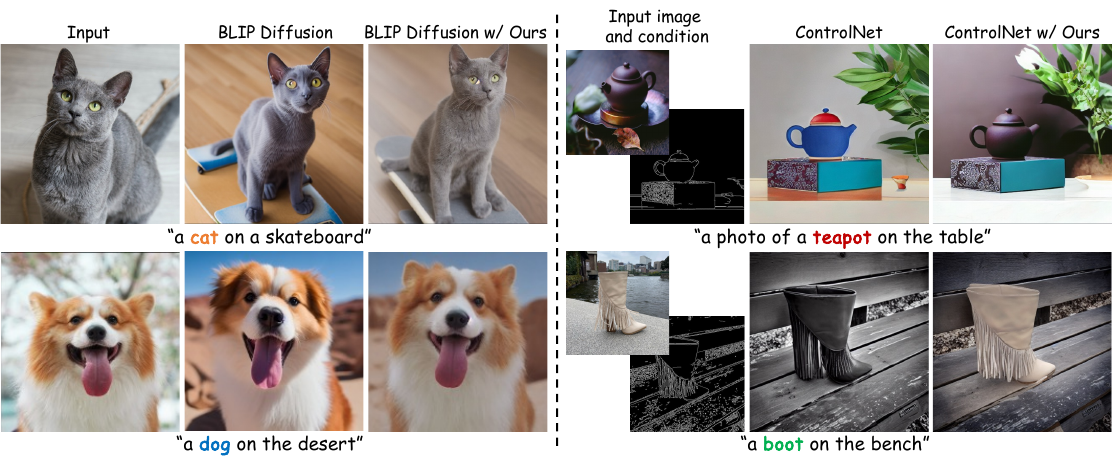}
  \caption{
  \textbf{Combined with existing methods.} (Left) Combined with our methods, BLIP Diffusion generates images more faithful to the input image. (Right) Using our approach, ControlNet yields results consistent with the input condition layout and the appearance of the input image.
  }
  \label{fig:combined_with_blip_diffusion_and_contronet}
\end{figure}

\noindent
{\bf Empower other methods.}
Our method can enhance ControlNet~\cite{zhang2023adding} and BLIP Diffusion~\cite{li2024blip} in a plug-and-play manner. As illustrated in Fig.~\ref{fig:combined_with_blip_diffusion_and_contronet}, by using our method, the output of BLIP diffusion becomes more faithful to the input image and better aligned with the input text. Furthermore, ControlNet can generate results that are consistent in layout and identity when combined with ours.

\begin{figure}[t]
  \centering
  \includegraphics[width=1\linewidth]{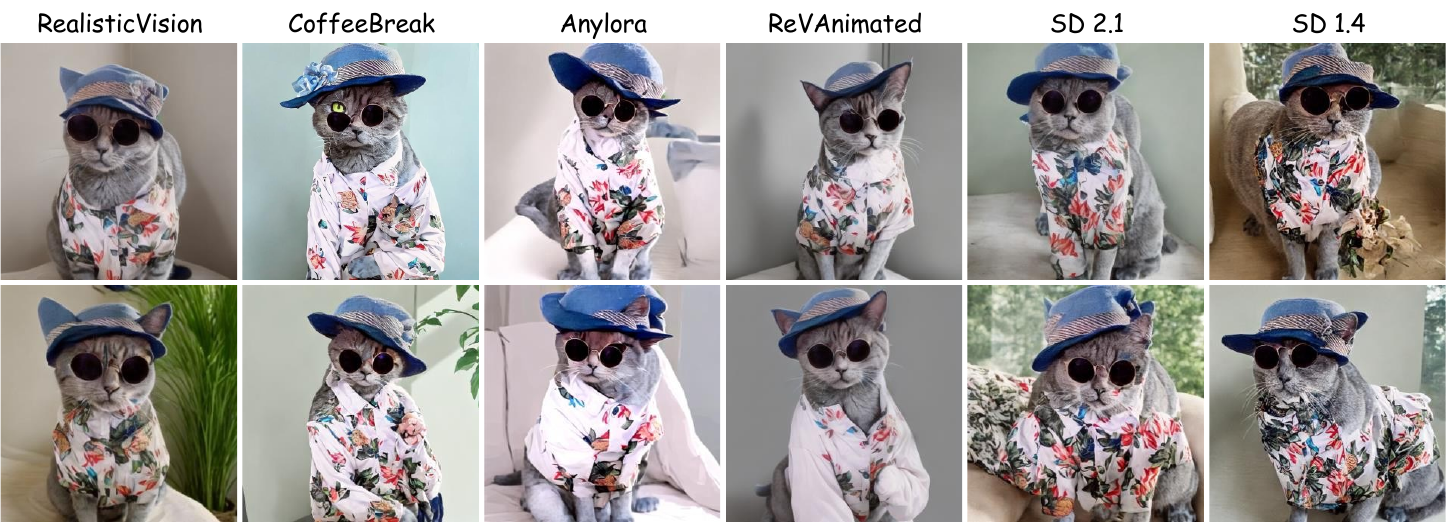}
  \caption{
  \textbf{Applying our method to different base models} with the same images and prompt as input. The input concepts including cat, hat, sunglasses, and floral shirt are well preserved.
  }
  \label{fig:apply_different_base_models}
\end{figure}

\noindent
{\bf Applied to different base models.}
Our method exhibits strong flexibility and robustness, as it can directly work on different base models. As illustrated in Fig. \ref{fig:apply_different_base_models}, we apply our method to various versions of the SD model and different checkpoints on the Civitai platform.
\section{Conclusion}\label{sec:conclusion}

In this paper, we propose a novel tuning-free method to address the challenges of multi-concept customized composition. Experimental results show that our proposed method enables flexible combinations of different objects from various categories. However, our method currently faces limitations in that it lacks an explicit module to perceive the structure of the input reference concepts. In our future work, we plan to integrate additional techniques such as image adapter, to overcome these limitations.

\textbf{Acknowledgement}
This work was in part supported by
National Key R\&D Program of China (No.\  2022ZD\-0118\-700).

{
\small
\bibliographystyle{ieeenat_fullname}
\bibliography{ref.bib}
}

\newpage
\appendix
\renewcommand\thefigure{A\arabic{figure}}
\renewcommand\thetable{A\arabic{table}}
\renewcommand\theequation{A\arabic{equation}}
\setcounter{equation}{0}
\setcounter{table}{0}
\setcounter{figure}{0}

\section*{Appendix}

\section{Paradigm Comparison}

Fig.~\ref{fig:paradigm_comparison} in the main paper visually represents the differences between our method and the other two types of methods. 

In the context of training-based methods, we can categorize them into two types, both of which take a set of 3-5 images representing the same concept as input. Type \uppercase\expandafter{\romannumeral1} methods, such as DreamBooth~\cite{ruiz2023dreambooth} and Break-A-Scene~\cite{avrahami2023break}, require training all the weights of the U-Net model. In contrast, type \uppercase\expandafter{\romannumeral2} methods, including Textual Inversion~\cite{gal2022image}, XTI~\cite{voynov2023p+}, and NeTI~\cite{10.1145/3618322}, typically focus on training a text-embedding that captures the semantic information of the input text, rather than training the entire network.

Furthermore, there is a second category of methods (BLIP Diffusion~\cite{li2024blip}, ELITE~\cite{DBLP:conf/iccv/WeiZJB0Z23}) tailored for generalizable customization. These methods typically involve two stages. In the first stage, the model is pre-trained on a large-scale dataset to learn the distribution within a specific domain. In the second stage, the pre-trained model is used in a training-free manner. However, this type of method is time-consuming, as it requires extensive pre-training, and applying it to a new model necessitates retraining.

In contrast to the aforementioned methods, our tuning-free approach differs in several key aspects. Firstly, it does not necessitate training the entire network or learning a text-embedding on a few images. Additionally, our method does not rely on pre-training on a large-scale dataset. Instead, our approach only requires a single image per concept as input. By replacing the self-attention mechanism in select blocks of the U-Net network with MRSA (Multi-Reference Self-Attention), our method enables single-concept customization and multi-concept composition. This means that our method can be easily applied to different models.

In summary, our method stands out by offering a more efficient approach to customization compared to the other two types of methods. It leverages MRSA to enable flexible customization of single or multiple concepts. 

\section{More Visual Results}

\subsection{Single-concept Customization}
\begin{figure*}[t]
  \centering   
  \includegraphics[width=1\linewidth]{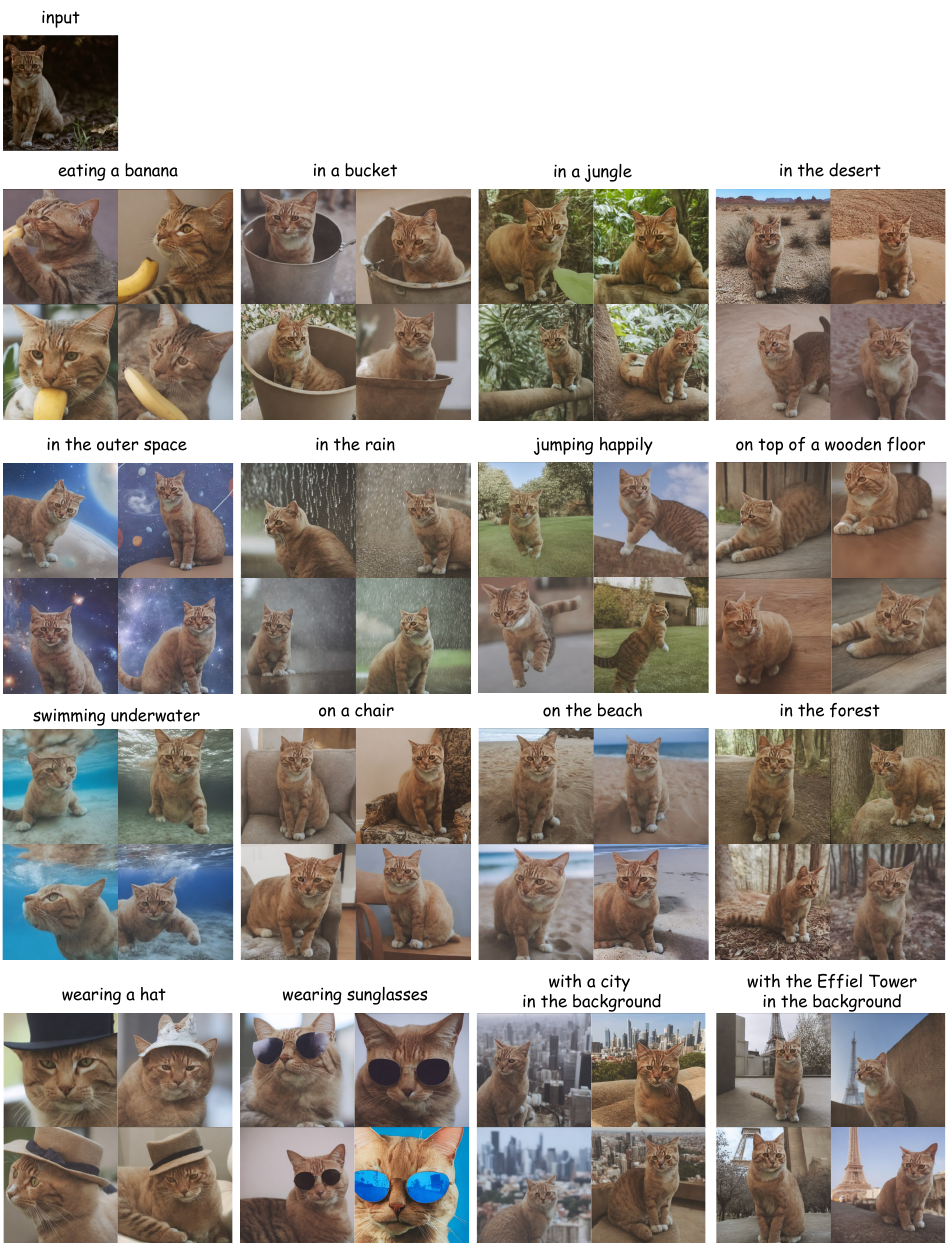}
  \caption{
  \textbf{Single concept qualitative results.} Our method enables extensive customization of a single concept by inputting a single image. 
  }
  \label{fig:cat1_qualitative_results}
\end{figure*}

Our approach allows for generating a diverse range of customized images based on a single concept in the input image. Fig.~\ref{fig:cat1_qualitative_results} showcases the results obtained using different seeds and target prompts. These results demonstrate the high fidelity and preservation of identity, highlighting the effectiveness and robustness of our approach.

\subsection{Empower Other Methods}
\begin{figure*}[t]
  \centering   
  \includegraphics[width=0.7\linewidth]{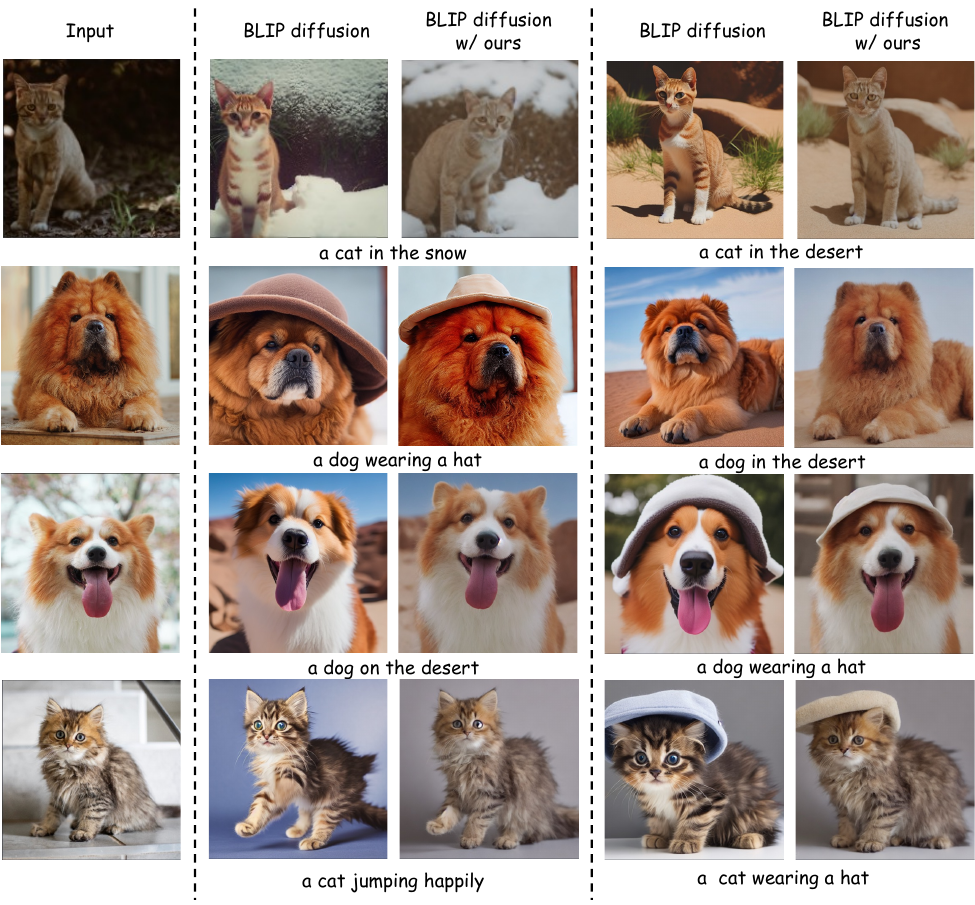}
  \caption{
  \textbf{BLIP diffusion vs. BLIP diffusion with FreeCustom.} The images generated by BLIP diffusion are visually appealing, but they may not achieve perfect identity recovery. However, when our approach is combined with BLIP diffusion, the generated results effectively maintain the identity of the given concept.
  }
  \label{fig:blipdiffusion with ours}
\end{figure*}

\begin{figure*}[t]
  \centering 
  \includegraphics[width=0.7\linewidth]{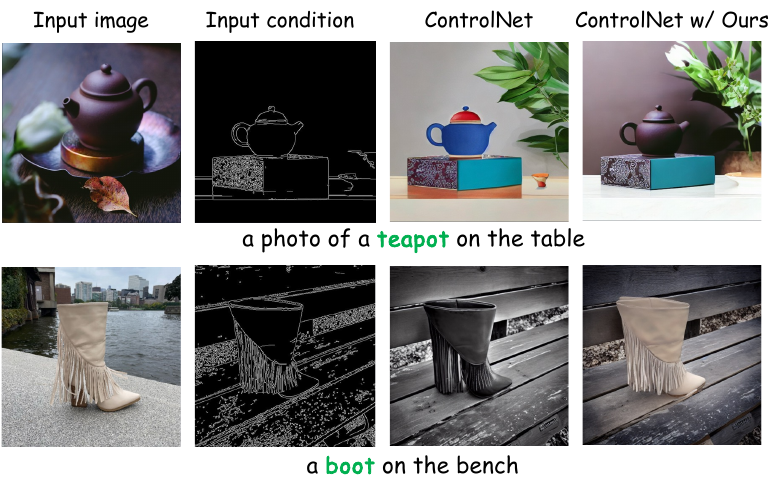}
  \caption{
  \textbf{ControlNet with FreeCustom.} Our method can be seamlessly integrate with ControlNet. From left to right, the images are arranged as follows: the input image, the input conditions, the image output by vanilla ControlNet, and the image output by ControlNet enhanced by FreeCustom.
  }
  \label{fig:ControlNet with ours}
\end{figure*}

Our methods have the capability to enhance other methods in terms of identity preservation and visual fidelity. When combined with BLIP diffusion, our methods effectively maintain the identity of the given concept as shown in Fig.~\ref{fig:blipdiffusion with ours}. Moreover, our method can be seamlessly integrated with ControlNet, as illustrated in Fig.~\ref{fig:ControlNet with ours}.

\section{Correspondence Visualization}
\begin{figure*}[t]
  \centering 
  \includegraphics[width=1\linewidth]{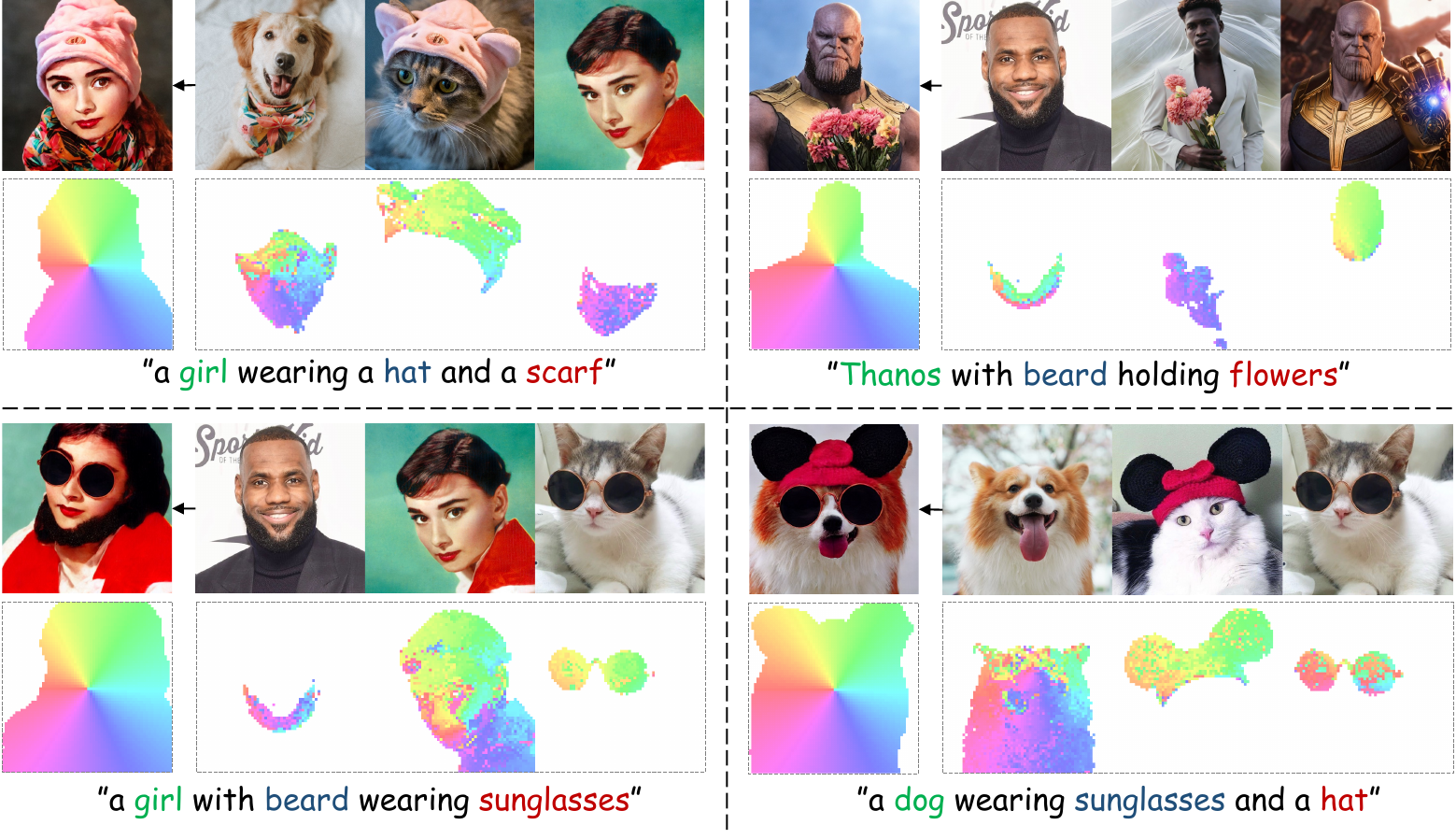}
  \caption{
  \textbf{Correspondence visualization.} We visualize the correspondence between each feature in the reference concepts and each feature in the generated multi-concept composition image using an attention map. In this figure, the features with the highest similarity between the combined concept and the reference concepts are marked with the same color to indicate the correspondence between them. The results indicate that we have achieved relatively good consistency across the features. For instance, the hat and scarf in the reference concept exhibit a strong match with the hat and scarf in the generated image, underscoring the effectiveness of our approach.
  }
  \label{fig:corr}
\end{figure*}

To illuminate the feature-wise correspondence between the reference concepts and the generated multi-concept composition image, we employ a visualization based on an attention map from layer 10 during the 50th denoising step. As depicted in Fig.~\ref{fig:corr}, pixels exhibiting the highest similarity between the combined and reference concepts are marked with matching colors on the map.

\section{Visualization of Multi-attention Map}
\begin{figure*}[t]
  \centering 
  \includegraphics[width=1\linewidth]{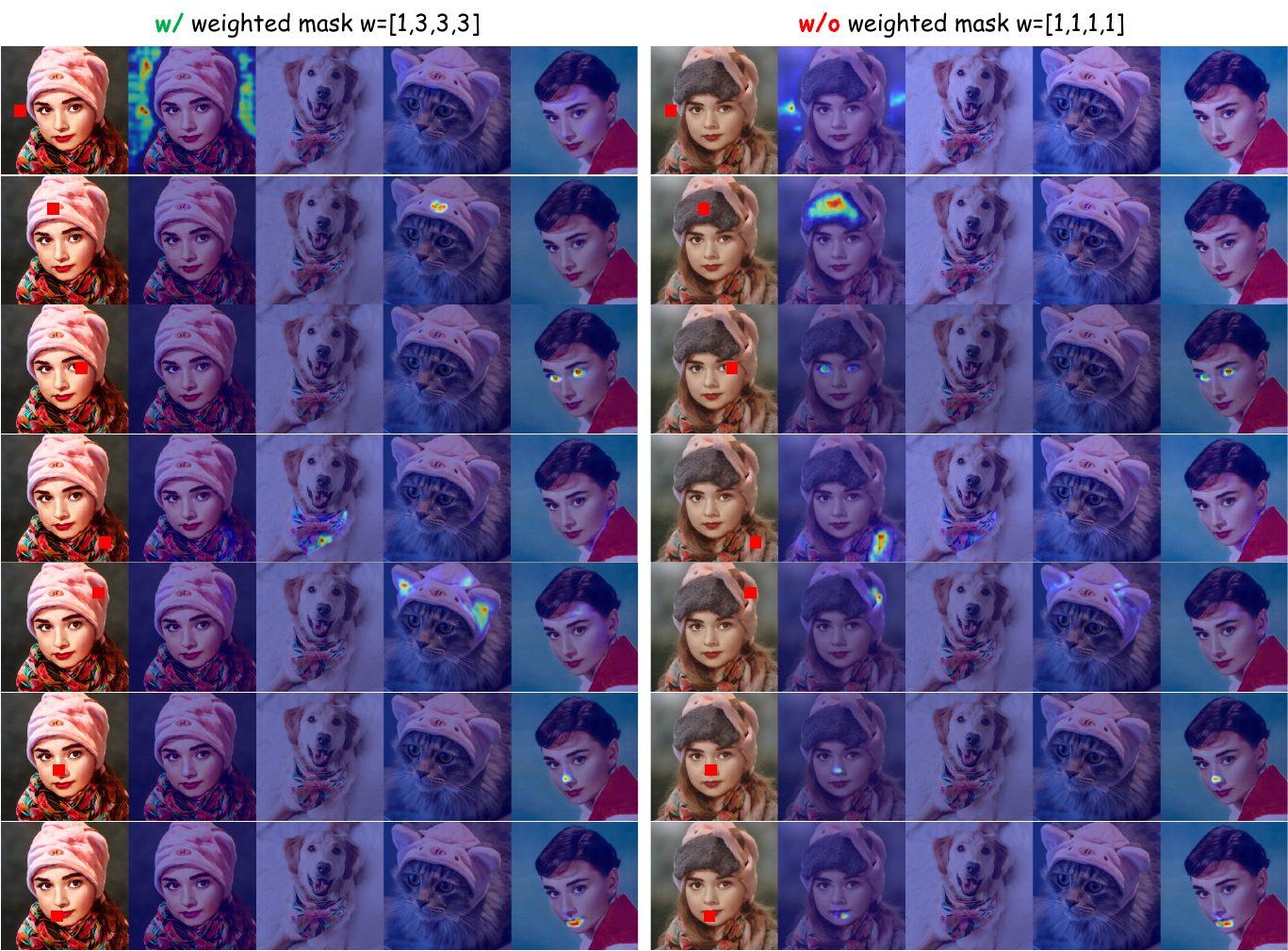}
  \caption{
  \textbf{Visualization of multi-attention map.} Each column of pictures in the figure represents, from left to right: the image output by our method (1st column), and the multi-attention map $A=\operatorname{Softmax} (\frac{\mathbf{M}_w \odot (\mathbf{Q} \mathbf{K}'^T)}{\sqrt{d}})$, $A\in{\mathbb{R}^{H\times ((N+1)W)}}$, $N=3$ here (the following 4 columns). The red box indicates the feature that queries other keys.  When using the weighted mask, the generated image will prioritize the reference concepts, while without the weighted mask, the output image will prioritize itself, resulting in a weaker preservation of the identity of the reference concepts.
  }
  \label{fig:attn_map_analyse_detail}
\end{figure*}

\begin{figure*}[t]
  \centering 
  \includegraphics[width=1\linewidth]{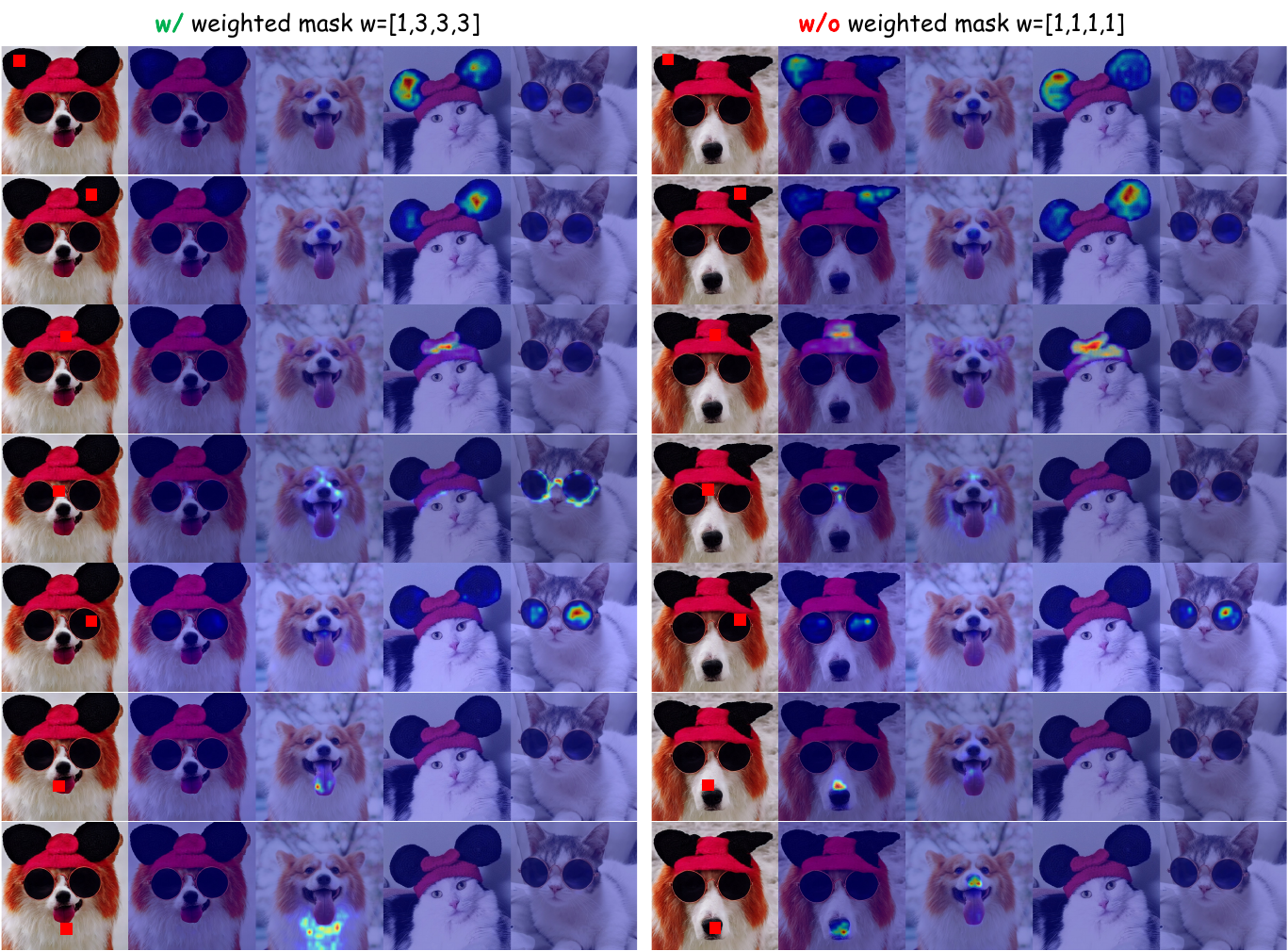}
  \caption{
  \textbf{Visualization of multi-attention map.} Each column of pictures in the figure represents, from left to right: the image output by our method (1st column), and the multi-attention map $A=\operatorname{Softmax} (\frac{\mathbf{M}_w \odot (\mathbf{Q} \mathbf{K}'^T)}{\sqrt{d}})$, $A\in{\mathbb{R}^{H\times ((N+1)W)}}$, $N=3$ here (the following 4 columns). The red box indicates the feature that queries other keys.  When using the weighted mask, the generated image will prioritize the reference concepts, while without the weighted mask, the output image will prioritize itself, resulting in a weaker preservation of the identity of the reference concepts.
  }
  \label{fig:attn_map_analyse_detail_2}
\end{figure*}

The visualization of the weighted mask's importance is depicted in Figs.~\ref{fig:attn_map_analyse_detail} and~\ref{fig:attn_map_analyse_detail_2}. In the left column, when using the weighted mask, we set $\mathbf w = [1,3,3,3]$. The generated image gives higher priority to the reference concepts. This is evident from the attention map, which shows that the image's attention is focused on the reference concepts.

In contrast, the right column represents the scenario without the weighted mask, where $\mathbf w = [1,1,1,1]$. Here, the output image prioritizes itself, with the attention being more concentrated on itself rather than the reference concepts. Consequently, this can lead to a weaker preservation of the identity of the reference concepts, such as the scarf and hat.

\begin{figure*}[t]
  \centering 
  \includegraphics[width=1\linewidth]{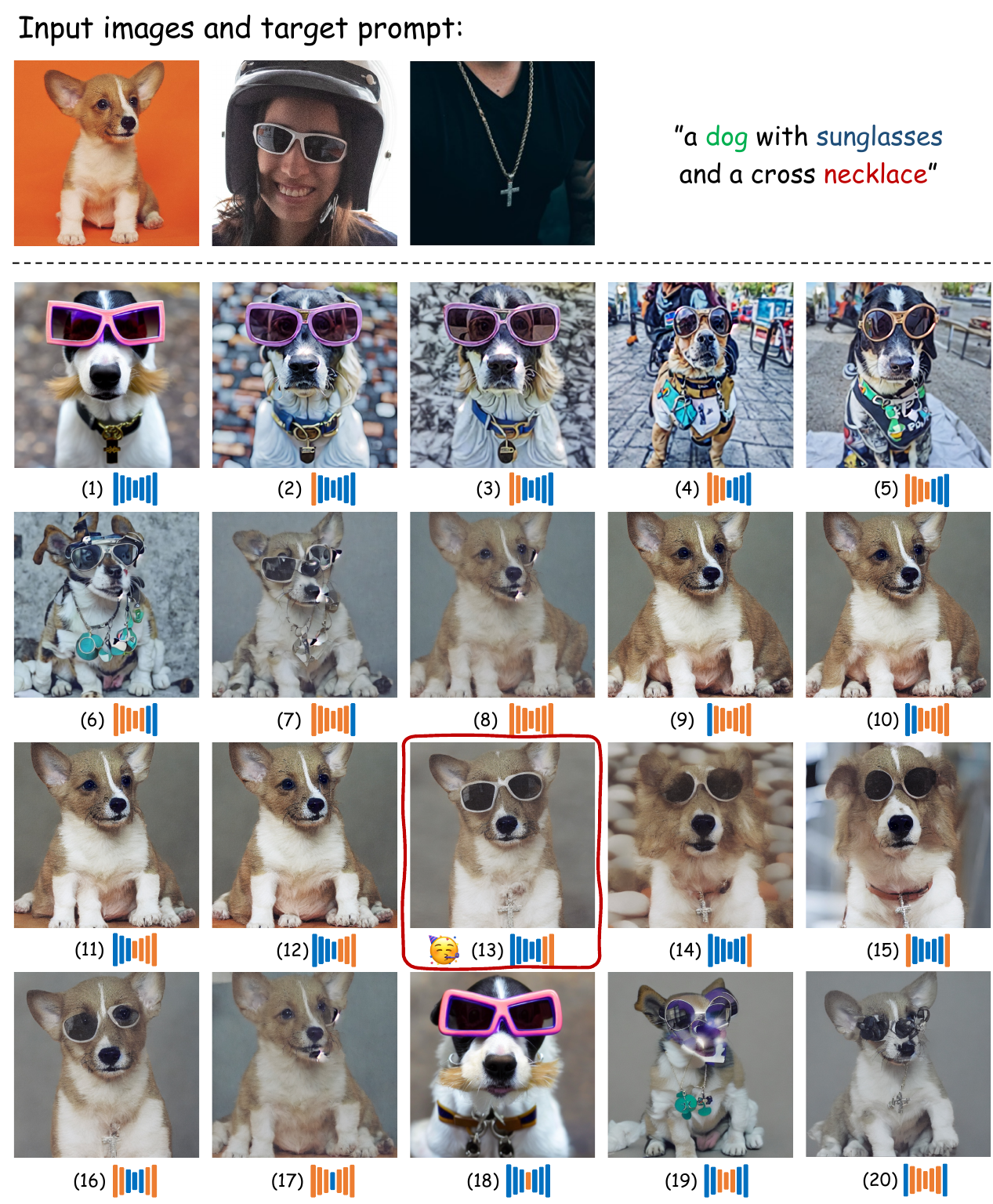}
  \caption{
  \textbf {Selective applying MRSA to basic blocks.} The blue color represents the original basic block and the yellow color indicates the basic block whose self-attention is replaced by MRSA.
  }
  \label{fig:block_ablation_supp}
\end{figure*}

\begin{figure*}[t]
  \centering 
  \includegraphics[width=1\linewidth]{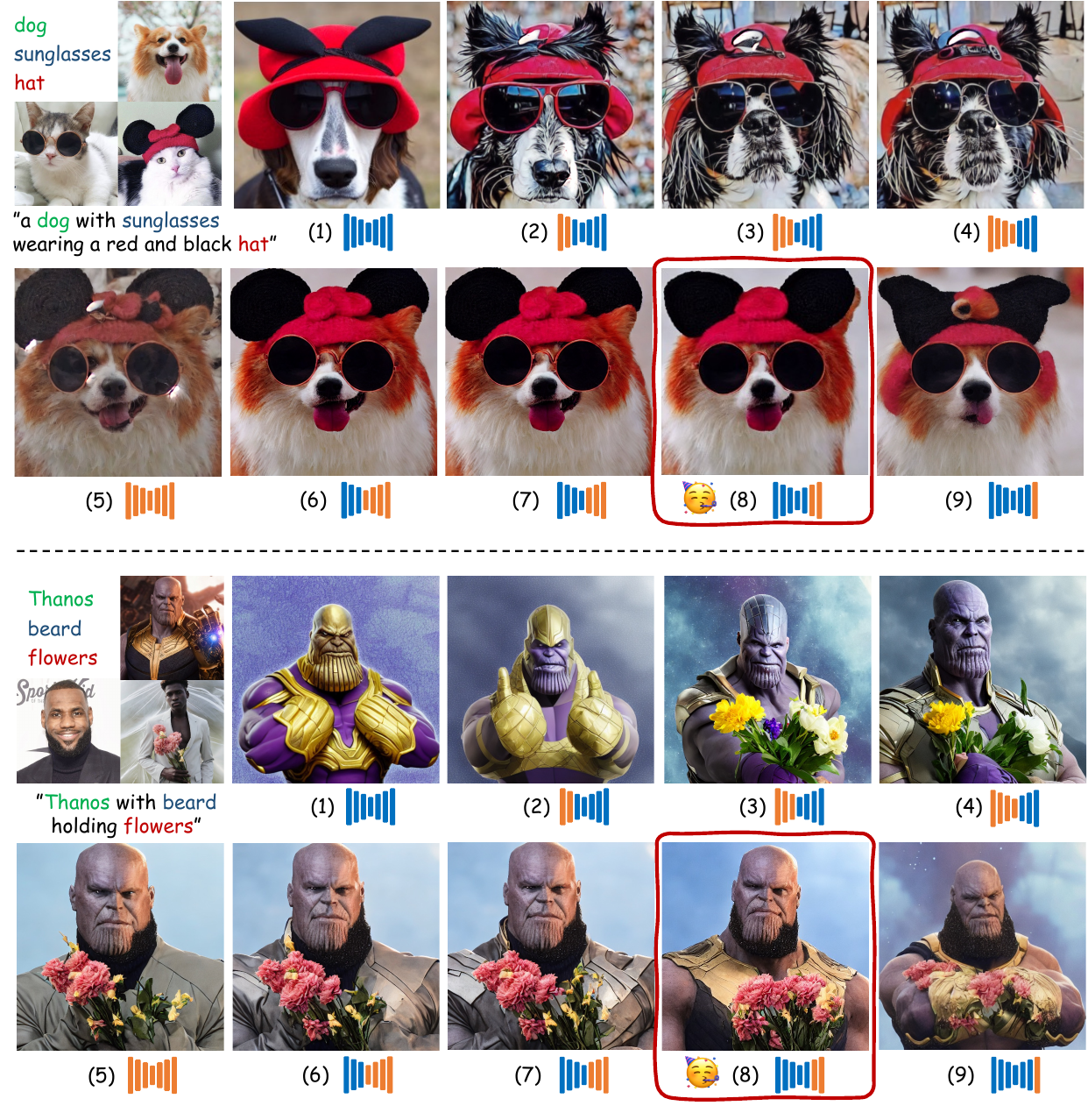}
  \caption{
  \textbf {Selective applying MRSA to basic blocks.} The blue color represents the original basic block and the yellow color indicates the basic block whose self-attention is replaced by MRSA.
  }
  \label{fig:block_ablation_supp_2}
\end{figure*}

\section{Selective MRSA Replacement}
We provide more comprehensive results of selectively replacing the original self-attention in the basic block with MRSA as shown in Fig.~\ref{fig:block_ablation_supp} and Fig.~\ref{fig:block_ablation_supp_2}. Fig.~\ref{fig:block_ablation_supp} shows a more fine-grained replacement strategy, Fig.~\ref{fig:block_ablation_supp_2} presents the ablation results of the selective replacement strategy when using different combinations of reference concepts as input.

\end{document}